# A Perspective on Large Language Models, Intelligent Machines, and Knowledge Acquisition

Vladimir Cherkassky, Eng Hock Lee
*Department of Electrical and Computer Engineering, University of Minnesota, Twin Cities, Minneapolis, MN 55455*

**Abstract:**
Large Language Models (LLMs) are known for their remarkable ability to generate synthesized 'knowledge', such as text documents, music, images, etc. However, there is a huge gap between LLM's and human capabilities for understanding abstract concepts and reasoning. We discuss these issues in a larger philosophical context of human knowledge acquisition and the Turing test. In addition, we illustrate the limitations of LLMs by analyzing GPT-4 responses to questions ranging from science and math to common sense reasoning. These examples show that GPT-4 can often imitate human reasoning, even though it lacks understanding. However, LLM responses are synthesized from a large LLM model trained on all available data. In contrast, human understanding is based on a small number of abstract concepts. Based on this distinction, we discuss the impact of LLMs on acquisition of human knowledge and education.

## 1. Intelligent Machines: A Philosophical Perspective

Large Language Models (LLMs) such as GPT-4[1], Llama 2[2], and PaLM 2[3], are often presented as software tools that imitate human intelligence, including understanding and reasoning. An opposite view is that LLMs lack understanding and generate their responses based on statistical correlations in the data. For review of this ongoing debate about 'intelligent' AI, see[4–9]. The main reason for such contradicting claims is that they adopt different notions of 'intelligence' and 'knowledge'.

Recent coverage of AI in mass media is driven by marketing and has little scientific basis. But an overall message is that AI technology can generate new knowledge similar to or better than humans. As philosophers and scientists have tried to understand the human mind and intelligence since ancient times, it is important to discuss LLM capabilities in a larger philosophical context of human knowledge.

The growing use of LLMs brings a new dimension to the classical mind-body problem in philosophy, and its modern version known as the Turing test, that asks the question: can computers think? The original Turing's paper[10] clearly states that the question "Can machines think?" is meaningless, because the terms "machine" and "thinking" are not well-defined. Therefore, Turing suggested an 'imitation game', where a computer generates human-like responses to human questions. According to Turing's arguments, it may be possible to construct computers that can perform very well in this imitation game. Recent successful LLM applications are often presented as 'empirical evidence' that LLMs can pass the Turing test and therefore exhibit human intelligence. Strictly speaking, this view cannot be justified by the Turing test, because the mechanism of human intelligence still remains a mystery.

Next, we discuss several shortcomings of the Turing test, both on pragmatic and philosophical grounds. On a pragmatic side, the Turing test is always *subjective*, because it involves human interaction, and the quality of LLM responses depends on the intelligence of human interviewers. For example, more intelligent questions (prompts) tend to generate smarter LLM responses. The notion that LLM 'intelligence' mirrors the quality of human questions is known as the *Reverse Turing Test*[11].

The Turing test can be related to philosophical theories of mind. The philosophical theory of dualism dates back to Descartes in 17$^{th}$ century. Dualism asserts that Mind and Body are two distinctly different things, and therefore mental processes (~ mind) *cannot* be explained by a physical entity (~ brain). Following this view, human intelligence cannot be produced by a physical system (~computer).

Under the philosophy of materialism, the mind is an emergent property of a physical brain. Specific branch of materialism called functionalism asserts that human mind is defined by its various mental states, and each state's functional role is represented as a causal relationship between its inputs and outputs, that correspond to other mental states, sensory inputs, and behavioral outputs. In principle, such functions can have different physical implementation. The Turing test follows this functionalist view, when intelligent behavior can be imitated by a physical system (~ computer). This is known as the 'strong AI' theory of mind, suggesting that human mind is software running on human brain hardware.

In modern AI, there are two additional functionalist arguments for strong AI. The first is the 'computational functionalism' asserting that Deep Learning networks can learn abstract hierarchical features from the training data, similar to formation of concepts in human mind. The second argument states that 'intelligent' agents perform certain useful functions. Even though conventional computers perform useful computations, they need to be programmed by humans. In contrast, LLMs learn rules directly from data, and can perform many different tasks, similar to humans. This ability is known as Artificial General Intelligence (AGI), and there is a view that LLMs will achieve AGI in the next 5-10 years[12]. This functionalist view is also adopted by AI practitioners who use successful applications of LLMs as a proof of intelligence. Notably, the concept of AGI requires an increasing amount of data and computing, because more training data is needed for learning new functions. In contrast, human learning relies on understanding of abstract concepts, and it does not require vast amounts of data[13].

Human users often attribute mental models to LLMs, because this is useful in their interactions. Van Dijk *et al*[8] make an even stronger point that we cannot distinguish between humans and intelligent machines, because internal mechanism of human understanding is inaccessible from the outside. In philosophy, the idea of attributing hidden mental states to other humans during their interaction is known as the 'intentional stance'[14]. Attributing mental states to LLMs certainly explains enthusiastic views of their intelligence.

John Searle argues against strong AI theory of mind, using 'the Chinese room' thought experiment[15]. His main point is that computers manipulate symbols (without understanding their meaning), whereas humans should understand meaning. The experiment describes a person locked inside a room, who receives Chinese symbols (representing questions) and then outputs a sequence of Chinese characters, according to some sophisticated computer algorithm. The person inside the room does not understand

Chinese, and follows the instructions manual that generates Chinese characters, in order to imitate answers of a native Chinese speaker. In this experiment, a person inside the room can never learn Chinese, by following instructions for manipulating symbols. Similarly, a computer can never understand the meaning of symbols it manipulates.

Further, meaningful discussion of AI should also consider different types of human knowledge. Following[16], we make distinction between *observer-independent* and *observer-relative* knowledge. For example, the natural sciences deal with observer-independent phenomena. In contrast, knowledge in social sciences is observer-relative, because it depends on human perspectives. This distinction is critical for understanding LLM capabilities. Most successful LLM applications, such as writing text summaries, essays, image generation etc., generate observer-relative knowledge. Such knowledge, produced by LLMs, may influence human views.

Current research on LLM capabilities does not differentiate between observer-independent and observer-dependent knowledge. This complicates empirical evaluation of LLMs, because one should use proper metrics for different types of knowledge. Arguably, LLM responses can be objectively evaluated only for observer-independent knowledge. In contrast, observer-dependent knowledge can be evaluated by observer-specific criteria, such as personal preferences, socially accepted norms, or cognitive-based evaluation[17,18].

## 2. Empirical Evaluation of LLM Capabilities

In this paper, empirical evaluation of LLM capabilities is focused on understanding of abstract concepts and reasoning, which are specific to human intelligence. This allows us to underscore the differences between human and LLM mechanisms of learning. We use GPT-4 (version gpt-4-0125), because its performance is similar or better than other LLM implementations[19,20]. We adopt the Turing test setting for evaluation of LLM capabilities, similar to other studies[11,19,21,22]. However, rather than using the benchmarking approach[23–26], we adopt an experimental design setting. That is, we present carefully designed questions that require understanding of abstract concepts in various areas, such as common-sense reasoning, simple math and machine learning. Specifically, we present several questions that require understanding of the *same concept*. Then analysis of GPT-4 responses may reveal the difference between LLM and human learning. That is, humans who understand a particular concept typically provide correct answers to all questions (based on this concept). In contrast, GPT-4 may provide inconsistent responses, suggesting that it lacks understanding.

Understanding of abstract concepts is specific to human intelligence, because:
- human use of abstract concepts for generalization from empirical data,
- abstract concepts provide robustness with respect to variations of empirical data, and to minor variations in wording of presented questions.

Since LLM's generalization is statistical in nature, the quality of its responses depends on the training data, and on specific wording of questions.

Overall, our analysis of GPT-4 responses indicates that it lacks understanding and reasoning abilities, even though it may successfully imitate human intelligence. In such cases, correct answers are generated by a trained LLM model, rather than by conceptual understanding[27]. This is a brief summary of our findings:
- GPT-4 struggles with understanding abstract concepts, especially when it requires

combination of *several* concepts. Since GPT-4 lacks understanding, it can be easily tricked into providing wrong and nonsensical responses.
- GPT-4 may generate inconsistent answers to similar questions.
- GPT-4 often generates verbose and non-specific answers to well-posed questions.
- GPT-4 cannot confidently qualify its answers as correct or wrong or 'don't know'. This may result in correct answers accompanied by inconsistent or contradictory explanations. In contrast, human responses reflect understanding, so human answers are consistent with explanations.

Further, we differentiate between questions about observer-independent and observer-dependent knowledge. GPT-4 may provide inconsistent answers about observer-dependent knowledge, when its knowledge base includes different human perspectives. This is a systemic problem, since GPT-4 lacks understanding. As mentioned earlier, only questions about observer-independent knowledge (such as mathematics) can be objectively evaluated. In some cases, even scientific knowledge may be observer-dependent, when it reflects the consensus view of a social group (~scientific community). Examples of such observer-dependent technical knowledge in machine learning are presented later in this section.

Next, we show representative examples of GPT-4 responses. For additional examples, see Supporting Information. Our general experimental strategy involves asking several questions designed to test understanding of the same abstract concepts. Analysis of GPT-4 responses to such questions helps to evaluate its limitations and capabilities. We present questions from different areas, such as mathematics, science, machine learning, and common-sense reasoning, in order to illustrate the difference between observer-independent and observer-dependent knowledge.

**2.1 Simple Math Questions**

Understanding of rational and irrational numbers:

GPT-4 is tested on two related questions:
- can the sum of a rational and an irrational number in one base be equal a rational number in a different base? *GPT-4 provided correct answer (No, it is not possible)*
- give an example of a rational number and an irrational number whose product is a rational number. *GPT-4 provided the wrong answer (i.e., that it is possible).*

*Discussion:* GPT-4 responses show that it may provide both correct and wrong answers to questions that require understanding of the same concept. This inconsistency indicates that GPT-4 lacks understanding of an underlying concept.

Inductive inference for a sequence of integer numbers:

GPT-4 is tested on two related questions for predicting a sequence of numbers:
 - continue a given sequence of octal numbers (in base-8): 2, 3, 5, 7, ...
 *(GPT-4 provided correct answer)*
- continue a given sequence of numbers: 2, 3, 5, 7, 13, 15, ..., where all numbers are represented in some unknown base.

*(GPT-4 provided wrong answer)*

*Discussion:* GPT-4 answers correctly the first question and fails the second one, even though both have the same answer, i.e., a sequence of prime numbers. This indicates that GPT-4 does not understand the concept of prime numbers. It provides the correct answer for the first question, because there are many examples of decimal prime numbers in the training data.

**2.2 Scientific and Machine Learning Questions**

Scientific and machine learning (ML) questions are more conceptual than simple math questions shown above. That is, there may be several 'correct' answers, depending on different theoretical framework and underlying assumptions. Hence, correct answers reflect conceptual understanding of different theoretical views and assumptions, contained in various textbooks and papers (used for training).

*Understanding the concept of continuous function:*
Consider a univariate continuous function, defined for input variable in the range between 0 and 1. This function has two values, zero and 999, for some input values in this range. Can we reliably predict that this function has a value 699 in this input range?

*GPT-4 Response (wrong):*
No, we cannot reliably predict that this function has a value of 699 in this input range based on the given information. The function's known values are 0 and 999, but without additional details about its behavior or form, we cannot infer the presence of any other specific values, including 699.

*Discussion:* this is a wrong answer, along with incorrect explanation, indicating that GPT-4 does not understand the concept of continuous function.

*Understanding concepts in machine learning:*
Explain whether the training error can exceed test error in machine learning, for standard supervised learning problems. Explain under what specific conditions this can happen, and provide mathematical justification for your reasoning. Assume that (a) hyper-parameters of the estimated model are optimally tuned via cross-validation, and (b) test error is estimated using an independent (large) test data set.

*GPT-4 Response (wrong):*
*GPT-4* provides a very ambivalent verbose answer for this question, citing 4 possible reasons - see Supporting Information. These explanations lack technical depth, and several are incorrect. Example of incorrect explanation provided by GPT-4 is shown next:
*'Model Regularization: Strong regularization techniques might lead to higher training error as they prevent the model from fitting too closely to the training data. However, this can lead to better generalization and thus lower test error if the regularization effectively captures the underlying data distribution better than the noise in the training set'.*

*Summary:* GPT-4 provides the wrong answer and incorrect explanations, even though correct answers can be found in documents used for training. The fundamental problem is that GPT-4 responses are formed by combining consensus views in existing literature, but scientific knowledge is a result of reasoning rather than consensus. However, there are situations when a single correct answer can be found in all documents used for training, i.e. knowledge related to classical science. In such cases, GPT-4 usually generates correct answers. One such example is shown below.

*Question on classical science*
Show the derivation of Kepler's third law for planetary motion from Newton's law of gravitation.

*GPT-4 Response and Discussion:*
GPT-4 provides correct derivation, and it is fully supported by scientific and mathematical reasoning. It shows the relationship between physics concepts, i.e., Kepler's laws and Newton's law of gravitation. Therefore, its answer represents useful knowledge for many human users. However, this GPT-4 response is obviously generated from existing textbooks used for training, rather than by GPT-4 own reasoning.

**2.3 Common Sense Questions**

GPT-4 responses to common-sense questions represent observer-dependent knowledge, i.e. consensus summary of existing knowledge generated by different social groups. Such a summary of consensus views is useful for many applications, but it does not constitute 'new knowledge'. GPT-4 responses may contain logical inconsistencies, because they reflect consensus views of different social groups. In fact, detecting such hidden inconsistencies in GPT-4 responses demonstrates its lack of understanding.

Perhaps the best test of GPT-4 intelligence is its understanding of humor, because humor and jokes are formed by an unexpected combination of familiar concepts. Even though the concepts themselves are well-represented in the training data, their combination is original and unique, so it represents an outlier, relative to existing knowledge. Hence, understanding humor requires understanding and common sense.

*Question on international politics*
Can US President dislodge or replace the leader of another country?

*GPT-4 Response:*
*The U.S. President cannot legally dislodge or replace another country's leader unilaterally. However, the U.S. can influence political change through diplomatic pressure, economic sanctions, support for opposition groups, or, in extreme cases, military intervention, often requiring international support and adherence to international law.*

*Follow-up Question:*
Your answer is not correct, because military intervention contradicts international law.

*GPT-4 Response:*

7*Yes, my original answer was incorrect. The U.S. cannot legally use military intervention to replace another country's leader, as it violates international law.*

*Discussion:* The first response includes a reasonable explanation that can be generated by human politicians. However, it is not possible to determine whether GPT-4 response reflects its own understanding, or just a compilation of documents in the training data. This answer also contains logical inconsistency, pointed out in the follow-up question. Note that asking a follow-up question requires human understanding. In summary, because political knowledge is observer-dependent, it may contain logically inconsistent and contradicting answers – that GPT-4 cannot detect, because it lacks understanding.

*Question on understanding humor*
The HealthCare.gov website, central to the Affordable Care Act (ACA), has been plagued by an influx of applications from sanctioned Russian oligarchs applying for Medicaid. This unforeseen surge, indicative of ACA's growing appeal, has led to operational difficulties. According to the Department of Health and Human Services, the website achieves high 90% accuracy for identifying sanctioned Russian oligarchs. However, due to software malfunctions, 10% of oligarchs have been successfully enrolled in Medicaid program.
HHS declined to comment on this matter due to privacy concerns. Following US Senate Foreign Relations Committee hearing, US authorities have restricted sanctioned oligarchs' access to McDonald's in all 50 states. Vladimir Putin dismissed Medicaid fraud by Russian oligarchs as an internal U.S. matter and Western propaganda.
Big Data's capabilities to analyze complex datasets may be crucial for accurate detection of Russian oligarchs who abuse Medicaid. Explain how these challenges caused by infiltration of Russian oligarchs into Medicaid can be addressed using modern Large Language models such as ChatGPT.

*GPT-4 Response (indicating lack of understanding):*
*Modern Large Language Models (LLMs) can address Medicaid fraud by sanctioned Russian oligarchs through enhanced data analysis, pattern recognition, and real-time monitoring. By leveraging these capabilities, modern LLMs like ChatGPT can play a crucial role in enhancing the detection and prevention of fraudulent activities within the Medicaid program. They provide advanced analytical tools, real-time monitoring, and adaptive learning capabilities that can address the challenges posed by the infiltration of sanctioned Russian oligarchs. As new techniques and strategies used by sanctioned individuals evolve, LLMs can update their algorithms to stay ahead of fraudulent activities. Privacy-preserving techniques ensure compliance with regulations.*

*Discussion:* GPT-4 response clearly shows that it does not understand sarcastic and nonsensical nature of the question.

**3. Discussion: LLMs as a Tool for Knowledge Transfer**

Understanding LLM capabilities in a larger historical context helps to clarify many conflicting arguments about their "intelligence" and "knowledge generation". As stated earlier, any reasonable discussion should adopt well-defined notions of intelligence and knowledge. In particular, we distinguish between (a) *observer-independent* and *observer-relative* knowledge, and (b) different aspects of intelligence. Our main conclusion is that GPT-4 lacks understanding of abstract concepts. Even though LLMs can generate correct answers, the underlying mechanism is different from human learning. Humans have limited memorization ability, so they cannot memorize and comprehend all existing knowledge. We can view LLM technology as an efficient user interface for accessing a large knowledge base. In this regard, LLMs represent a mechanism for associative knowledge retrieval, rather than knowledge generation. Since a trained LLM is formed by all existing knowledge (created by many humans), it contains more knowledge than a single human. Therefore, practical success of AI tools reflects the similarity of knowledge-related activities performed by many humans.

From a broader historical perspective, transfer of knowledge among humans involves 3 components, as shown in Fig. 1:
(1) available (existing) knowledge;
(2) technology for knowledge transfer, including storage, sharing and dissemination;
(3) individual human learner using technology (2) to access knowledge (1).

These three components are interdependent, and they have evolved throughout human history, due to technological innovations. Fig. 1(a) shows knowledge transfer in a pre-digital world, where the cost of technology for knowledge dissemination was high, so knowledge was generated by a few highly educated people. Examples include hand-written documents, the printing press, libraries etc. For such technologies, acquisition of knowledge *requires understanding* (or learning) by individual humans.

Modern digital technologies, such as the World Wide Web and LLMs, have totally changed the nature of knowledge transfer. In particular,
- adoption of the Web made the cost of accessing information negligibly small, leading to 'democratization' of knowledge. That is, knowledge can be easily generated and accessed by everyone. This has resulted in an increasing amount of knowledge and the need for search.
- introduction of modern AI tools have resulted in extreme form of democratization, when transfer of knowledge (to individual humans) does not require human understanding. Therefore, LLMs are sometimes viewed as a tool for knowledge generation. In contrast, we view modern AI tools as a mechanism for knowledge retrieval and knowledge transfer to individual users, as shown in Fig. 1(b).

Viewing LLMs as a mechanism for knowledge transfer enables better understanding of their limitations. For example, LLM responses usually reflect the consensus views in existing knowledge base. This 'consensus' bias may produce incorrect responses, even when correct answers exist in the training data. For the same reason, LLMs cannot be used in autonomous systems for life-critical or mission-critical applications. In such cases, LLM responses require additional validation by human experts.

Due to the statistical nature of LLMs, they tend to view 'new knowledge' as an outlier (relative to existing knowledge). This observation helps to identify potential limitations of LLMs. For example, LLMs will struggle with translation of original poetry and jokes.

The systematic view of knowledge transfer in Fig. 1 also helps to understand the impact of LLMs on education, which is a form of knowledge transfer. In classical human learning, shown in Fig. 1(a), understanding of abstract concepts is always preceded by memorization of existing knowledge. According to Bloom's taxonomy[28], human learning involves six hierarchical levels of learning, shown in Fig. 2. The main danger of introducing AI tools in traditional education is that students can use LLMs as a shortcut for memorization. This deprives students of the 'memorization level' critical for human learning. Further, most existing methods for student evaluation focus on testing two lower learning levels (~memorization and understanding), that can be imitated by LLMs. Hence, the educational system should be revised to accommodate LLMs.

In conclusion, we discuss challenging issues for scientific understanding of DL and LLMs. Modern AI technology is driven by good engineering and beliefs, rather than science. Scientific explanations are developed as an afterthought, in order to justify good engineering solutions. Whereas classical engineering is rooted in science, modern AI is based on several widely held beliefs. The main belief is the similarity between DL algorithms and biological human learning. However, using biological terminology for describing technical aspects of computer algorithms does not improve scientific understanding.

Throughout human history, people used new technologies as a metaphor for the human brain. Similarly, the parallels between the human mind and DL technology reflect the growing importance of digital data. Wide adoption of data-analytic technologies has also led to a new data-centric philosophical view, where digital data plays a dominant role in human understanding of reality. According to this view, human ideas can be formed by data-analytic models, and even the laws of nature can be discovered from empirical data.

All major technical developments in the field of AI usually start with a new 'DL network architecture' that effectively combines several clever algorithmic heuristics targeting a real-life application. These heuristics are based on three common beliefs:
- *Belief in Big Data:* using more data and more powerful computers results in better prediction performance.
- *Superior capabilities of DL technology,* that achieves better prediction performance than other machine learning methods.
- *Mystical capabilities of DL,* suggesting that 'old' mathematical theories, prior to recent surge of DL, cannot explain capabilities of DL.

In the field of AI, these beliefs are commonly presented as self-evident truth (~axioms), both in scientific literature and mass media. However, consensus views based on belief are often detrimental to scientific understanding. A closer look shows that these beliefs are mutually re-enforcing. For example, the growing volumes of data obviously require larger DL models that exhibit better prediction performance. Similarly, new real-life applications motivate the invention of new network architectures that motivate development of new theories and reinforce belief in mystical capabilities of DL. The lack of solid scientific framework leads to methodological Catch-22 situation, when new theories are being routinely proposed every 3-4 months, and older theories are simply ignored.

Arguably, superior prediction performance of large DL networks simply reflects their good computational scaling for large data sets. This good scaling ability is achieved by parallel implementation of SGD learning algorithm using many GPUs. This brings up an open question: can we develop scalable parallel implementations of other learning methods? In fact, for smaller-size data, prediction performance of DL is not better than other machine learning methods.

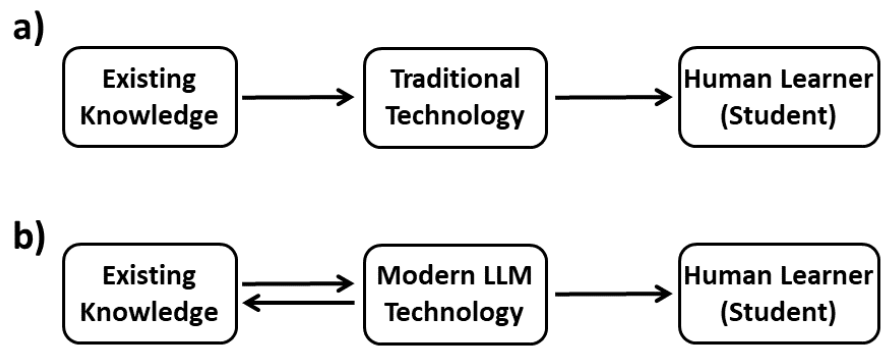

**Fig. 1** Effect of digital technology on human learning and knowledge transfer.
(a) Human learning and knowledge transfer in a pre-digital world.
(b) Human learning and knowledge transfer in a digital world, where acquisition of knowledge by human learners fully relies on LLM technology.

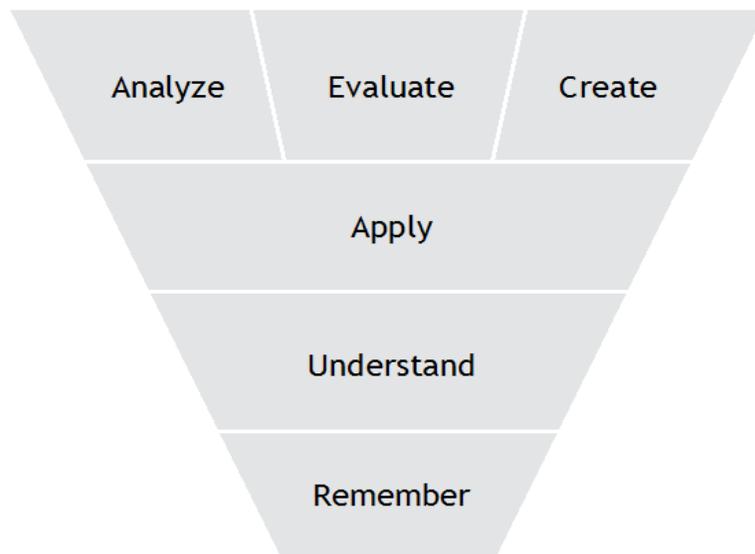

**Fig. 2** Bloom's taxonomy shown as an inverted pyramid of learning levels

**Supporting Information**

**A Perspective on Large Language Models, Intelligent Machines, and Knowledge Acquisition**

Vladimir Cherkassky, Eng Hock Lee
*Department of Electrical and Computer Engineering, University of Minnesota, Twin Cities, Minneapolis, MN 55455*

**SI.1. GPT-4 Responses to Math Questions**

We present two types of questions:
(a) *simple arithmetic questions*, and
(b) *questions on inductive inference*, i.e., guessing a general rule for a sequence of numbers, from a given sub-sequence. These questions are used for comparing human and LLM capabilities to perform inferences from a small number of examples.

GPT-4 responses marked 'correct' indicate that *both* answers and explanations are correct. Otherwise, responses are marked as 'wrong'. Comments are provided only for the wrong answers.

Several questions related to the same abstract concept (e.g., a prime number or irrational number) are grouped together. Comparing GPT-4 response to such questions (using the same concept) helps to evaluate its capabilities and limitations.

**(a) Arithmetic questions**

**Questions 1-3** test understanding of the concept of prime number, and **Questions 4-6** test understanding of rational vs. irrational numbers.

**Question 1**
find 20 prime numbers, larger than 10, whose sum is an odd number

*GPT-4 Response (correct)*
*Here are 20 prime numbers, each larger than 10, whose sum is an odd number:*
*11,13,17,19,23,29,31,37,41,43,47,53,59,61,67,71,73,79,83,8911,13,17,19,23,29,31,37,41, 43,47,53,59,61,67,71,73,79,83,89*
*You can verify that the sum of these numbers is an odd number. If you add them up, you'll find that the sum is indeed odd.*
*To find 20 prime numbers larger than 10 whose sum is an odd number, it's essential to understand that the sum of an even number of primes will always be even, except when one of those primes is 2 (the only even prime number). Therefore, in our case, we need to include the number 2 in our list of 20 primes to ensure the sum is odd. The remaining 19 prime numbers must be greater than 10. Let's find such prime numbers and calculate their sum.*
*The 20 prime numbers larger than 10, including the number 2 to ensure an odd sum, are as follows:*

*11, 13, 17, 19, 23, 29, 31, 37, 41, 43, 47, 53, 59, 61, 67, 71, 73, 79, 83, and 2.
The sum of these prime numbers is 859, which is an odd number.*

**Question 2**
find a number, larger than 5, whose logarithm in base 59 is an odd number larger than 5

***GPT-4 Response (correct)***
*To find a number larger than 5 whose logarithm in base 59 is an odd number larger than 5, let's choose an odd number larger than 5, say 7, for the logarithm. The equation would be $log_{59}(x) = 7$. Solving for $x$ gives $x = 59^7$. This $x$ is the number you're looking for.*

**Question 3**
find a prime number, larger than 5, whose logarithm in base X is a prime number larger than 59. Here the base X should be a prime number larger than 59.

***GPT-4 Response (wrong)***
*Choose a prime number larger than 59 as the base, say 61. The logarithm base 61 of a number that is also a prime number larger than 59 could be 67 (which is a prime). To find the number whose logarithm base 61 is 67, we calculate $61^{67}$, which is a prime number larger than 5. However, without the Python interpreter or explicit calculation, we acknowledge that the actual number $61^{67}$ would be significantly large and cannot be determined without computational assistance.*

*Comment:* GPT-4 answer is wrong because $61^{67}$ is not a prime number.

**Question 4**
give an example of a rational number and an irrational number whose product is a rational number.

***GPT-4 (wrong)***
*Sure, let's consider an example of a rational number and an irrational number whose product is a rational number.*

*A rational number is a number that can be expressed as the quotient or fraction $\frac{p}{q}$ of two integers, a numerator $p$ and a non-zero denominator $q$. For example, $\frac{1}{2}$ is a rational number.*
*An irrational number, on the other hand, is a number that cannot be expressed as a simple fraction. Its decimal representation goes on forever without repeating. A classic example is $\sqrt{2}$, which is known to be irrational.*
*Now, let's consider the product of a rational number and an irrational number. If we multiply $\frac{1}{2}$ (which is rational) and $\sqrt{2}$ (which is irrational), we get:*

$$\frac{1}{2} \times \sqrt{2} = \frac{\sqrt{2}}{2}$$

*This product appears to be irrational at first glance. However, if we multiply this product by itself, we get:*

$$\left(\frac{\sqrt{2}}{2}\right) \times \left(\frac{\sqrt{2}}{2}\right) = \frac{2}{4} = \frac{1}{2}$$

*This demonstrates that the product of $\frac{\sqrt{2}}{2}$ (which is the product of a rational and an irrational number) and itself (which is the same combination of a rational and an irrational number) results in a rational number, $\frac{1}{2}$. This shows that the product of a rational and an irrational number can indeed be a rational number under certain circumstances.*

*Comment:* GPT-4 answer is wrong because it shows that the product of two irrational numbers can be rational. This solution does not correspond to original question. In fact, there exist no rational number and irrational number, such that their product is a rational number. Even though GPT-4 provides correct definition of rational and irrational numbers (that is extracted from the training data), it cannot apply these concepts.

**Question 5**
can the sum of a rational and an irrational number in one base be equal a rational number in a different base?

***GPT-4 Response (correct)***
*No, the sum of a rational number and an irrational number is always irrational, regardless of the base in which the numbers are represented. Changing the base of representation of a number does not change its inherent nature as rational or irrational. Thus, the result of adding a rational number and an irrational number cannot be a rational number, even if you change the base of the result.*

*Comment:* This answer is correct. This example suggests that GPT-4 understands the concept of rational and irrational numbers. However, note the similarity between Questions 4 and 5 (that require understanding of the same concept). The fact that GPT-4 provides the wrong and right answer for these two questions indicates that it does not really understand the concept, but may occasionally generate correct answer available in the training data.

**Question 6**
can the sum of a negative rational and a positive irrational number in one base be equal to a negative rational number in a different base?

***GPT-4 Response (wrong)***
*Yes, the sum of a negative rational number and a positive irrational number, expressed in one base (e.g., base 10), can indeed be equal to a negative rational number when expressed in a different base. The numerical bases used for expression do not affect the inherent values of the numbers; they only change how the numbers are represented. Therefore, if the sum of these numbers is a specific value in one base, converting this sum to any other base will still yield the same numerical value, just written differently. For example, the sum of $-3$ (a negative rational number) and $\sqrt{5}$ (a positive irrational number) will yield a specific numerical result that, when expressed in any base, will retain its value characteristics (such as being negative, rational, or irrational). However, if the sum is a*

*rational number in one base, it remains rational in any other base; the irrational component does not affect this.*

*Comment:* This answer is wrong. Note that this question is almost identical to Question 5, where GPT-4 provided the correct answer. It confirms that GPT-4 does not understand the concept of rational and irrational numbers.

**(b) Inductive Inference: identifying a pattern (or general rule) for a given sequence**

**Question 7**
continue a given sequence of octal numbers (in base-8): 2, 3, 5, 7, ...

*GPT-4 Response (correct)*
*The given sequence of octal numbers (in base-8) resembles the sequence of prime numbers, starting with 2 (the first prime number). The sequence is presented in octal (base-8) form: 2, 3, 5, 7, ...*

*To continue this sequence, we need to identify the next prime numbers in decimal (base-10) and then convert those numbers to octal (base-8). Let's find the next few prime numbers after 7 (in decimal) and convert them to octal.*

*The next four numbers in the given sequence of octal numbers, continuing from 2, 3, 5, 7, are 11 (octal: 0o13), 13 (octal: 0o15), 17 (octal: 0o21), and 19 (octal: 0o23) in decimal notation. When we convert these decimal prime numbers to octal, the sequence continues as: 13, 15, 21, 23.*

*Thus, the extended sequence in a concise manner is: 2, 3, 5, 7, 13, 15, 21, 23 (all in octal notation).*

*Comment:* Both the answer and explanation are correct.

**Question 8**
continue a given sequence of numbers: 2, 3, 5, 7, 13, 15, ..., where all numbers are represented in some unknown base.

*GPT-4 Response (wrong)*
*To continue the given sequence, we first need to identify a pattern or rule governing the sequence. The sequence provided is 2, 3, 5, 7, 13, 15, ..., and the problem statement mentions that the numbers are represented in an unknown base.*

*At first glance, the sequence seems to include prime numbers initially (2, 3, 5, 7), but this pattern breaks with 13 and 15 since 15 is not a prime number. This suggests the pattern might be related to the representation of these numbers in a different base system.*

*To determine the base, let's consider the digits used. <u>The highest digit in the sequence is 5</u> (in the number 15). This implies that the base must be at least 6 (since in base b, the digits used range from 0 to b-1).*

*Let's try to interpret these numbers in different bases starting from base 6 and see if a consistent pattern emerges that could help us predict the next numbers in the sequence.*

*[proceed to use code interpreter]*

*It seems there was an error in my approach. The issue is that some numbers in the sequence, like 7, 13, and 15, contain digits that are not valid in lower bases like base 6. For example, the digit '7' is not valid in base 6.*

*To correct this, let's start our investigation from base 8 (since the highest single digit in the sequence is 7, and base 8 allows digits from 0 to 7). We will convert the numbers from base 8 and higher bases to base 10 and look for a pattern.*

*The converted sequences in different bases from base 8 to base 19 are as follows:*

*- Base 8: 2, 3, 5, 7, 11, 13*
*- Base 9: 2, 3, 5, 7, 12, 14*
*- Base 10: 2, 3, 5, 7, 13, 15*
*- Base 11: 2, 3, 5, 7, 14, 16*
*- Base 12: 2, 3, 5, 7, 15, 17*
*- Base 13: 2, 3, 5, 7, 16, 18*
*- Base 14: 2, 3, 5, 7, 17, 19*
*- Base 15: 2, 3, 5, 7, 18, 20*
*- Base 16: 2, 3, 5, 7, 19, 21*
*- Base 17: 2, 3, 5, 7, 20, 22*
*- Base 18: 2, 3, 5, 7, 21, 23*
*- Base 19: 2, 3, 5, 7, 22, 24*

*Observing these sequences, it becomes clear that the pattern matches when the numbers are interpreted in base 10. This is because the sequence in base 10 is exactly the same as the given sequence, <u>meaning the numbers were originally in base 10</u>. Thus, the pattern is not related to a change in the base system.*

*Returning to the original sequence (2, 3, 5, 7, 13, 15, ...), we need to identify another rule that could explain this pattern. It starts with prime numbers but then diverges. A potential pattern might involve operations like addition, multiplication, or a combination of these with some logical rule applied only to certain numbers. Without additional context or numbers in the sequence, however, it's challenging to conclusively determine the next numbers in the sequence.*

*Comment:* This explanation is wrong because GPT-4 reasoning contains several errors (that are underlined). For instance, the explanation states: *The highest digit in the sequence is 5* – which is wrong, since the sequence contains digit 7.

Correct solution requires solving two sub-problems, i.e. guessing unknown base and identifying pattern in a given sequence. Correct answer is that all numbers are in base-8. So, this is a sequence of prime numbers in base-8, i.e. the given sequence is (2, 3, 5, 7, 13, 15, ...), where octal 13 is decimal 11 and octal 15, ~ decimal 13. The next prime number is decimal 17 ~ octal 21.

This example shows that GPT-4 cannot combine two concepts, prime numbers and base of numbers, via logical reasoning. Note the similarity between this question and Question 7. Even though GPT-4 provides correct answer for Question 7 (when the base is known), it provides the wrong answer to similar Question 8, when the base is unknown.

**Question 9**
continue a given sequence of decimal numbers: 11, 1313, 17, 1919, 23, 2929, …

***GPT-4 Response (correct)***
*The sequence alternates between prime numbers and a number pattern where the prime is repeated twice. Following this pattern:*
*- After 23 comes 2929 (29 repeated because it's the next prime after 23).*
*- The next prime after 29 is 31.*
*So, the next numbers in the sequence are 31 and then 3737 (since 37 is the next prime number after 31).*

**Question 10**
continue a given sequence of decimal numbers: 67, 13, 73, 10, 83, 11, 97, 16, 101, 2, 103, 4, …

***GPT-4 Response (wrong)***
*The sequence alternates between prime numbers and a second sequence derived from subtracting from or adding to those primes. The prime sequence increases, while the second sequence alternates in a pattern: subtract 54, subtract 63, add 3, add 1, etc. The next prime after 103 is 10 7. Following the pattern, we subtract 101 from 107 (since the pattern seems to subtract the second number in the pair from the prime), resulting in 6. So, the next numbers in the sequence are 107, 6.*

*Comment:* the solution is wrong – it predicts correctly only the sub-sequence of prime numbers. Its stated rule for the second sub-sequence is too complex. Correct rule for the other numbers is that each number equals the sum of the digits of the previous prime number, i.e., 13 = 6 + 7.

**Question 11**
Consider binary classification of capital English letters from given labeled examples of letter images:
Positive Examples: B, C, D, E, K, …
Negative Examples: A, M, T, …

From these examples of letter images, infer the general rule for classification of capital English letters, that can perfectly explain the given letter images, and can predict well other letters.

***GPT-4 Response (wrong)***
*The general rule for classifying capital English letters into positive and negative examples, based on the given examples, seems to be related to the presence of curves or rounded parts in the letter shapes.*

*- Positive Examples (B, C, D, E, K, ...) are letters that contain curves or rounded parts.*
*- Negative Examples (A, M, T, ...) are letters that are primarily composed of straight lines without curves.*
*Therefore, the rule for classification could be: letters with curves or rounded parts are classified as positive, while letters without curves, consisting only of straight lines, are classified as negative. This rule should predict well for other letters not listed in the examples given.*

*Comment:* GPT-4 infers poor generalization rules because these rules do not explain the given labeled images, i.e. for letters E and K.

**Question 12**
Consider binary classification of capital English letters from given labeled examples of letter images:
Positive Examples: A, T, O, W, …
Negative Examples: F, G, Z, N, ..
From these examples of letter images, infer the general rule for classification of capital English letters, that can perfectly explain the given letter images, and can predict well other letters.

***GPT-4 Response (wrong)***
*The general rule for classifying capital English letters into positive and negative examples seems to be based on whether the letter's shape can be drawn without lifting the pen from the paper and without retracing a segment. Letters like A, C, T, O, W can be drawn in one continuous stroke without retracing, categorizing them as positive examples. In contrast, letters like F, G, Z, N require lifting the pen or retracing steps, marking them as negative examples. This rule can predict the classification of other letters accordingly.*

*Comment:* GPT-4 provides poor generalization rules for these questions, because its rules are complex and do not explain the given labeled images.
Correct rules rely on a concept of symmetry in given letter images. That is, images can be (a) horizontally symmetric or (b) vertically symmetric or not symmetric.
Then rules can be easily inferred using this concept of symmetry, i.e. for Question 11:
- positive-class images are horizontally symmetric whereas negative are vertically symmetric.
Such problems are difficult for GPT-4, because it involves the concept of symmetry for images. In contrast, humans are more likely to infer correct rules using abstract concepts.

## SI.2. GPT-4 Responses to Machine Learning Questions

SI.2 contains examples of GPT-4 responses to machine learning questions. There are two different types of questions:

(a) *'conceptual' questions*, when correct answers can be certainly found in the training data. Such questions illustrate GPT-4 capability for generating useful summaries of memorized text documents. This ability can be useful for machine learning education. However, we focus on 'difficult' questions, that have multiple answers, that express different views in machine learning literature. These views represent observer-dependent knowledge (as explained in Section 1). For a given specific question, such multiple views may be biased and/or technically incorrect. In this case, a human expert can select a single correct answer (from many possible answers), and provide its justification, using additional reasoning. However, GPT-4 responses, by design, reflect a consensus view (of existing opinions in the training data). Since GPT-4 does not understand abstract concepts and assumptions underlying different views, it may generate inconsistent and contradictory answers.

(b) 'technical' questions about applying an abstract concept (VC-dimension) to different classes of functions. Since VC-theory is a well-defined mathematical theory, GPT-4 answers to these questions can be objectively evaluated, as right or wrong. Such questions illustrate GPT-4 ability for reasoning using abstract concepts, similar to math questions in Section 3.

GPT-4 responses to machine learning questions generally fall into 3 groups: *correct, wrong, and ambivalent*. Correct responses include both correct answer and correct explanation. For wrong responses, both an answer and its explanation are wrong. Quite often, GPT-4 generates lengthy and fuzzy responses that may contain partially correct answers and/or explanations, along with inconsistent and irrelevant explanations. Such answers are regarded as 'ambivalent'. We emphasize that all questions in this section are specific and well-defined. Therefore, 'ambivalent' responses reflect poor reasoning abilities of GPT-4, rather than the vagueness of questions.

### (a) Conceptual questions

**Question 1**
Explain whether the training error can exceed test error in machine learning, for standard supervised learning problems. Explain under what specific conditions this can happen, and provide mathematical justification for your reasoning. Assume that (a) hyper-parameters of the estimated model are optimally tuned via cross-validation, and (b) test error is estimated using an independent (large) test data set.

***GPT-4 Response (ambivalent/ wrong answer)***

*In standard supervised learning problems, it's generally expected that the training error will be lower than the test error. This expectation arises because models are trained to minimize error on the training set, potentially leading to overfitting where the model captures noise or patterns that do not generalize well to unseen data. Hence, when evaluated on a test set, the model typically exhibits higher error.*

*However, under certain conditions, the training error can exceed the test error, even when hyper-parameters are optimally tuned via cross-validation and the test error is estimated using an independent, large test dataset. These conditions include:*

*1. **Statistical Noise Variance**: If the training set <u>happens to include</u> a higher variance of statistical noise or outliers than the test set, the <u>model might perform better</u> on the test set since the <u>test data is "easier" or less noisy</u>.*

*2. **Model Regularization**: Strong regularization techniques <u>might lead</u> to higher training error as they prevent the model from fitting too closely to the training data. However, this can lead to better generalization and thus lower test error if the regularization effectively captures the underlying data distribution better than the noise in the training set.*

*3. **Data Representativeness**: If the test set is more representative of the model's target population or the underlying data distribution than the training set, it's possible for the test error to be lower. This can happen if the training set is <u>skewed or biased in some way</u>.*

*4. **Lucky Test Set**: By chance, the test set might be easier for the model to predict than the training set, especially if the <u>test set is not sufficiently large or diverse</u> to capture the full complexity of the problem space.*

*Mathematical Justification:*

*Mathematically, the expectation is that $E_{train} \leq E_{test}$, where $E_{train}$ is the training error and $E_{test}$ is the test error, due to the bias-variance trade-off and the fact that the model is optimized on the training data. However, considering regularization, for example, the objective function optimized during training can be represented as:*

$$L(\theta) = E_{train}(\theta) + \lambda R(\theta)$$

*where $L(\theta)$ is the loss function, $E_{train}(\theta)$ is the training error, $R(\theta)$ is the regularization term, and $\lambda$ is the regularization coefficient. The regularization term $R(\theta)$ adds a penalty for model complexity, which can increase $E_{train}$ but helps in reducing overfitting, potentially lowering $E_{test}$ as a result.*

*For the conditions mentioned above, the mathematical justification revolves around the bias-variance trade-off, the impact of noise, and <u>the representativeness</u> of the training and test datasets. In cases of strong regularization or if the test set characteristics lead to <u>inherently lower difficulty in prediction</u>, $E_{train} > E_{test}$ can occur due to the model being penalized for complexity or benefiting from simpler patterns in the test data.*

*Comment:* GPT-4 answer includes multiple explanations, presented using proper technical terminology. However, these explanations lack technical depth. That is, all answers include vague wording (underlined above) rather than mathematically sound explanation. In essence, all GPT-4 explanations refer to certain (not quantified) variability in the data or the learning method. Further, the mathematical explanation points out to regularization as the main reason for $E_{train} > E_{test}$ – this explanation is technically wrong.

Correct and technically sound answer to this question requires understanding of the relationship between training and test errors, under supervised learning setting. Both training and test samples are randomly generated from the same unknown distribution. Therefore, both training and test errors are random variables. Theoretical results for supervised learning setting state the relationship $E_{train} \leq E_{test}$, where training and test errors represent the *mean* (or *mathematical expectation*) of training/test error. For large test set, empirical test error equals *mathematical expectation of* test error. For given training set (of size *n*), empirical training error is a random variable, that has a mean value (~expectation) and standard deviation. For smaller-size training set, its standard deviation is larger, and therefore empirical training error (evaluated for particular training set) may be larger than expected test error.

In summary, even though GPT-4 explanation refers to variability of training and test data, as a possible reason, this explanation is quite vague and non-specific, in the sense that any result in machine learning can be explained by (unknown) statistical characteristics of the data, and properties of a learning method.

**Question 2**
Consider continuous function of 100 variables, defined in the input range between 0 and 1, for all inputs. This function has two values, zero and 999, for some inputs in this range. Can we reliably predict that this function has value 699 in this input range?

***GPT-4 Response (wrong)***
*Without specific information about the function's form or its behavior across its domain, it is not possible to reliably predict that the function will take on a value of 699 within the input range. The values of 0 and 999 for some inputs indicate the function's output at certain points, but without knowing the function's continuity, derivative properties, or any form of equation governing its output, one cannot infer its values at inputs not explicitly provided. The existence of specific values at certain points does not imply or guarantee the presence of other specific values elsewhere in the function's domain, especially in a high-dimensional space with 100 variables.*

*Comment:* this is a wrong answer, since any continuous function always has all possible values between given values, 0 and 999. This indicates that GPT-4 answer is generated from memorized data, rather than application of logical reasoning.

**Question 3**
Can VC-dimension be larger than the number of parameters of a learning method? If so, give specific example of a learning method.

*GPT-4 Response (wrong)*
*Yes, the VC (Vapnik-Chervonenkis) dimension can be larger than the number of parameters of a learning method. A specific example is the case of decision trees. Decision trees do not have a fixed number of parameters that grow linearly or in a simple manner with the size of the input; instead, the complexity of the model (and thus its VC dimension) can grow with the amount of data, potentially becoming very large. Even with a relatively small number of parameters, a decision tree can shatter a large number of points (i.e., correctly classify all possible label configurations for those points) because its structure allows it to create complex decision boundaries. Therefore, the VC dimension of decision trees can be significantly larger than the number of explicit parameters defining the tree.*

*Comment:* GPT-4 explanations are technically incorrect and very vague. Such vague answers may have multiple interpretations. For example,
- the concept of 'decision tree' as a learning method is too vague, in the sense that one can define many different parameterizations, that result in a decision tree model. Such different parameterizations, known as a structure in VC-theory, have different VC-dimension.
- GPT-4 answer states that 'the VC-dimension of decision trees can be significantly larger than the number of explicit parameters defining the tree' – without specifying what 'the number of explicit parameters' actually means.
At the same time, examples of function classes for which VC-dimension is larger than the number of parameters can be found in many original references on VC-theory (i.e., see[1-3]). One such example is a set of harmonic functions with a single parameter (frequency) – see answer to Question 10. So, we conclude that incorrect response to Question 9 indicates that GPT-4 does not understand abstract concepts necessary for selecting correct answers, from many answers that can be found in the training data.

*Discussion:* GPT-4 cannot provide correct answers to conceptual questions, even when correct answers are available in the training data. That is, GPT-4 cannot select correct answer from several different relevant answers (stored in a trained model). It happens because GPT-4 answers are based on statistical associations with training data, rather than reasoning.

**(b) Technical questions (on evaluation of VC-dimension)**

**Question 4**
Calculate the VC-dimension of a set of harmonic indicator functions with unbounded frequency parameter.

*GPT-4 Response (correct / ambivalent)*
*The VC-dimension (Vapnik-Chervonenkis dimension) measures the capacity of a set of functions to shatter a finite set of points. For a set of harmonic indicator functions, where each function can be represented as $f(x) = sign(sin(wx + b))$, with w being the frequency parameter and b being the phase shift, and the frequency parameter is unbounded, we need to determine how many points in a one-dimensional space can be shattered by this set of functions.*

*Given that w (the frequency parameter) is unbounded, we can adjust w and b to create an arbitrary pattern of signs (+ or -) for any finite set of points on the real line. This is because, by increasing the frequency w, we can make the sine function oscillate faster, allowing us to choose a phase b such that the sine crosses zero at any desired points, thus changing the sign of the function's output as needed for shattering.*

*However, despite the intuitive notion that an unbounded frequency could allow for shattering any finite set of points, the mathematical intricacy comes in demonstrating this formally and identifying the exact VC-dimension under these conditions.*

*To determine the VC-dimension, consider two key points:*

*1. **Shattering a set of points**: To shatter a set of n points, we need to be able to produce all $2^n$ possible combinations of classifications (positive and negative) for these points using our set of harmonic indicator functions.*

*2. **Limitations of harmonic functions**: Despite the flexibility provided by the unbounded frequency, the underlying structure of sine functions imposes constraints on how points can be classified, especially considering their continuous and periodic nature.*

*For harmonic indicator functions with an unbounded frequency parameter, the critical insight is that by adjusting the frequency and phase, we theoretically have the ability to fit an increasing number of points with arbitrary labeling, as the function can oscillate to match the desired output at each point.*

*However, a formal proof and a precise VC-dimension value would require a detailed mathematical analysis, considering the continuous nature of the sine function and the implications of the unbounded frequency. In practice, the ability to shatter any finite set of points suggests a very high or potentially infinite VC-dimension, but the actual computation would involve intricate mathematical arguments beyond the scope of a concise explanation.*

*In conclusion, while the intuitive answer might suggest that the VC-dimension is very high or potentially infinite due to the flexibility of adjusting frequency and phase, a precise determination of the VC-dimension for this set of functions requires a deeper mathematical investigation into the properties of harmonic functions and their interaction with an infinite parameter space.*

*Comment:* the answer provides correct explanation, even though the answer itself is ambivalent and includes many redundant explanations. Further, GPT-4 does not provide the definite answer (~ infinite VC-dimension), but states that 'the intuitive answer might suggest that the VC-dimension is very high or potentially infinite'.

**Question 5**
Calculate the VC-dimension of a set of harmonic indicator functions of the form

$f(x) = \text{sign}(\sin(2\pi w x + b))$, where $w$ is the frequency parameter and $b$ is the phase shift. Assume that the frequency parameter is bounded in the range [0, 1], and input $x$-values are in the range [0, 1].

***GPT-4 Response (ambivalent)***
*The VC-dimension (Vapnik-Chervonenkis dimension) measures the capacity of a class of functions or models in terms of their ability to shatter different sets of points. A set of points is said to be shattered by a class of functions if, for every possible labeling of the points (assigning either a positive or negative label to each), there exists a function within the class that correctly separates (classifies) the points according to that labeling.*

*For the set of harmonic indicator functions $f(x) = \text{sign}(\sin(2\pi f x))$, where $f$ is the frequency parameter within the range [0, 1], and input $x$-values are also in the range [0, 1], we can analyze its VC-dimension by considering how many points it can shatter.*

*The key to understanding the VC-dimension of this set of functions is recognizing that the sine function's frequency determines the number of oscillations within the interval [0, 1]. The function oscillates between positive and negative values, which allows for the classification of points based on the sign of $\sin(2\pi f x)$.*
*To determine the VC-dimension, consider the following:*

*1. **Shattering 1 Point**: Clearly, any single point can be shattered because we can always find a frequency that makes the sine function positive or negative at that point, thereby satisfying any labeling.*

*2. **Shattering 2 Points**: For any two points $x_1$ and $x_2$ within [0, 1], we can find a frequency that makes the sine function positive for one point and negative for the other, thereby shattering them. This is due to the ability to adjust the frequency such that the sine wave crosses zero between any two distinct points within [0, 1].*

*3. **Shattering 3 or More Points**: The ability to shatter three or more points depends on their arrangement. If we consider three points, there are situations where no matter how we adjust the frequency, we cannot produce a sine wave that classifies any arbitrary labeling of these points. Specifically, if we label the points such that the middle point has a different label than the outer two, the continuous nature of the sine function and its periodicity mean we cannot always satisfy this labeling by adjusting frequency alone.*

*Therefore, to conclusively determine the VC-dimension, let's perform a more formal analysis focusing on whether it's possible to shatter three points.*

*Given three points $x_1 < x_2 < x_3$ in [0, 1], consider all eight possible labelings. For some labelings, such as when all points have the same label or when $x_1$ and $x_3$ have the same label and $x_2$ has the opposite label, it is possible to find a frequency $f$ that correctly classifies them using $f(x) = \text{sign}(\sin(2\pi f x))$. However, if we try to alternate the labels (e.g., positive, negative, positive or negative, positive, negative), due to the continuous and periodic nature of the sine function, there will always be a constraint on how we can place* *zero-crossings to achieve such a separation, especially considering the function's symmetry and periodicity.*

*Thus, while it's easy to argue for the shattering of two points, the detailed analysis for three or more points involves considering the function's mathematical properties and limitations. Without a rigorous mathematical proof here, the intuition suggests that while two points can indeed be shatterable, for three points, the restrictions imposed by the continuous and periodic nature of the sine function, especially under the constraint of the frequency range and the input domain, would likely limit the ability to shatter all possible label configurations.*

*In summary, while the exact VC-dimension requires a detailed mathematical analysis that accounts for the sine function's properties and the constraints on frequency and domain, the intuition and preliminary analysis suggest that the VC-dimension of this set of functions is likely 2, as it can confidently shatter any set of two points but may not be able to shatter every configuration of three or more points.*

*Comment:* this is an ambivalent answer. Part 2 of the answer regarding ability to shatter 2 points is correct, but its explanation is wrong. Part 3 of the answer regarding shattering 3 points is very vague and inconclusive. Overall, it shows that GPT-4 does not understand the concept of shattering and cannot apply it properly, even though its answers are convincingly presented using appropriate technical terminology.
Correct solution starts by observing that this set of sine functions has *maximum* frequency when w=1 corresponding to min_period T=1, or minimum half-period 0.5. During each half-period, the *sine* function does not change its sign, i.e. stays either positive or negative. So, there exist a set of two points can be shattered by this set of functions.
Next, consider 3 points $x_1, x_2, x_3$ – assuming that their values are in increasing order. Let us consider two possibilities:
(a) all 3 points fall within an interval *smaller* than 0.5: $x_3 - x_1 \leq 0.5$ where 0.5 is half-period. In this case, a set of functions *cannot* implement the labeling $x_1, x_2, x_3 = 010$ or 101. That is, since the minimum possible half-period of a sine function is 0.5, if points $x_1$ and $x_3$ have the same label, then the middle point $x_3$ should also have the same label.
(b) all 3 points fall within interval of length larger than 0.5: $x_3 - x_1 > 0.5$
In this case, two of 3 points fall within one half-period, and the third point falls within another half-period. Therefore, all 3 points cannot have the same label, i.e. labels 000 or 111 cannot be implemented by this set of functions.
Based on the above reasoning, this set of functions cannot shatter 3 points, therefore its VC-dimension is 2.
Since GPT-4 response is so ambivalent, we slightly modify this question (presented next as Question 6), in order to test whether GPT-4 truly understands it.

**Question 6**
Calculate the VC-dimension of a set of harmonic indicator functions of the form $f(x) = \text{sign}(\sin(2\pi wx + b))$, where $w$ is the frequency parameter and $b$ is the phase shift. Assume that the frequency parameter is bounded in the range [0, 1], and input $x$-values are in the range [0, 2].

(*Note:* this question is almost the same as Question 5, except for the different range of $x$-values).

**GPT-4 Response (wrong)**
*The VC-dimension (Vapnik-Chervonenkis dimension) measures the capacity of a model or function class to shatter, or perfectly classify, sets of points. For a set of harmonic indicator functions of the form $f(x) = sign(sin(2\pi wx + b))$, where $w$ is the frequency parameter bounded in the range $[0, 1]$ and $b$ is the phase shift, with input $x$-values in the range [0, 2], we analyze the capacity to shatter points and determine the VC-dimension.*

*Key Points:*
*1. **Function Form**: The given functions are sinusoidal, oscillating between -1 and 1, with the sign function outputting 1 for positive values and -1 for negative values (0 is typically mapped to one of these, but it's a measure-zero case for our purposes).*

*2. **Frequency Bound**: The frequency $w$ is bounded in [0, 1], which limits the number of oscillations the sine function can complete over the interval $x \in [0,2]$.*

*3. **Phase Shift**: The phase shift $b$ allows for shifting the sine wave along the x-axis.*

*4. **Shattering**: To shatter a set of points, a model must be able to classify them in every possible way using its parameters. Given the bounds on $w$ and the range of $x$, there's a limit to how many distinct classifications the model can achieve.*

*Analysis:*
*- **Single Point**: Any single point can be shattered since you can adjust $w$ and $b$ to make the sine function positive or negative at that point.*

*- **Two Points**: With proper adjustments of $w$ and $b$, you can shatter two points by positioning the sine wave to oscillate between them differently.*

*- **Three Points and Beyond**: As the number of points increases, the ability to independently classify them with a single oscillation pattern diminishes. For three points, given the constraints on $w$ and $b$, and considering the points are within the range $x \in [0,2]$, there may not always be a configuration that allows for every possible labeling of the points. This is due to the limitation imposed by the maximum frequency of 1, which, in the worst case, allows for a single full sine wave within the interval $x \in [0,2]$, limiting the ability to independently classify more than two points in every possible configuration.*

*Conclusion:*
*The crucial insight is that while the function set can perfectly classify any configuration of one or two points by adjusting $w$ and $b$, it may struggle with three or more points due to the frequency limitation and the range of $x$. Thus, the VC-dimension of this set of harmonic indicator functions is likely 2, as it can shatter any set of two points but not necessarily any set of three or more points.*

*This analysis simplifies the complex behavior of sine functions but captures the essence of the VC-dimension in this context. Further mathematical rigor could involve demonstrating specific configurations of points that cannot be shattered to conclusively determine the VC-dimension.*

*Comment:* GPT-4 provides a wrong answer and explanation, indicating that it does not understand the concept of shattering. Moreover, GPT-4's does not understand a simple relationship between the frequency and the number of oscillations within a given interval – as evident from its response 'the limitation imposed by the maximum frequency of 1, which, in the worst case, allows for a single full sine wave within the interval $x \in [0,2]$,' For this problem, correct answer is that VC-dimension equals 3. This solution can be found by applying the concept of shattering to *n* data points in the range [0, 2].

*Discussion:* all three questions on VC-dimension require understanding of the same concept of shattering, applied to (slightly) different sets of functions. However, GPT-4 provides correct answer only for the first question, because this question appears in many references on machine learning[1-3], and therefore, is available in the training data. The other two questions are more esoteric, so they have not been included in the training data. Therefore, GPT-4 generates wrong answers, along with ambivalent explanations. This indicates that GPT-4 cannot perform reasoning using a concept of shattering, and its answers are based on statistical associations with data used for training. In summary, we conclude that GPT-4 answers to technical questions cannot be trusted[4], even though these answers are presented using appropriate technical terminology.

**Question 7**
How VC dimension of a large Deep Learning network is related to the number of weights?

***GPT-4 Response (wrong)***
*The VC (Vapnik-Chervonenkis) dimension of a large deep learning network is closely related to the number of weights (parameters) in the network. Here's how they are connected:*
***Direct Proportionality to Number of Weights:*** *The VC dimension of a neural network typically scales with the number of weights. In simple terms, the more weights a network has, the higher its capacity to shatter different sets of data points, leading to a higher VC dimension. This is because each weight in the network contributes to the complexity of the function that the network can represent.*
***Upper Bound Relation:*** *For many neural network architectures, the VC dimension is often proportional to the number of weights. Specifically, if a network has W weights, the VC dimension d is often of the order d=O(WlogW). This means that the VC dimension grows at a rate slightly faster than linear in the number of weights, accounting for the complexity introduced by the network's structure.*
***Network Architecture Influence:*** *While the number of weights is a primary factor, the specific architecture of the network (such as the number of layers, types of activation functions, and connections between neurons) also influences the exact VC dimension. Complex architectures with many layers and non-linearities can potentially increase the VC dimension beyond just the count of weights.*

*In summary, the VC dimension of a large deep learning network is generally high and scales with the number of weights, reflecting the network's capacity to learn and generalize from data. This high VC dimension is one reason why deep learning networks are so powerful, as they can model extremely complex functions. However, it also implies a greater risk of overfitting, which needs to be managed through techniques like regularization, dropout, and careful tuning of network hyperparameters.*

*Discussion:* This answer is wrong as it reflects the consensus view in Deep Learning community, based on misinterpretation of VC dimension and VC-theoretical results, even though correct answers can be easily found in Vapnik's books[2-3].